\pdfoutput=1

\documentclass[11pt]{article}

\usepackage[]{acl}

\usepackage{times}
\usepackage{latexsym}

\usepackage[T1]{fontenc}

\usepackage[utf8]{inputenc}

\usepackage{microtype}

\usepackage{inconsolata}

\usepackage{pgffor}
\usepackage{amsmath}
\usepackage{cleveref} 
\usepackage{graphicx}
\usepackage{listings}
\usepackage[absolute,overlay]{textpos}
\title{Assessing Generalization for Subpopulation Representative Modeling\\via In-Context Learning}

\author{Gabriel Simmons \\
  University of California, Davis \\
  \texttt{gsimmons@ucdavis.edu} \\\And
  Vladislav Savinov \\
  Independent Researcher \\
  \texttt{vlad.al.savinov@gmail.com} \\}

\begin{document}
\maketitle
\begin{abstract}
This study evaluates the ability of Large Language Model (LLM)-based Subpopulation Representative Models (SRMs) to generalize from empirical data, utilizing in-context learning with data from the 2016 and 2020 American National Election Studies. We explore generalization across response variables and demographic subgroups. While conditioning with empirical data improves performance on the whole, the benefit of in-context learning varies considerably across demographics, sometimes hurting performance for one demographic while helping performance for others. The inequitable benefits of in-context learning for SRM present a challenge for practitioners implementing SRMs, and for decision-makers who might come to rely on them. Our work highlights a need for fine-grained benchmarks captured from diverse subpopulations that test not only fidelity but generalization.
\end{abstract}

\begin{textblock*}{\textwidth}(1in,\dimexpr\textheight+3.5cm\relax)
\footnotesize 
\noindent\rule{\textwidth}{0.4pt}\\
Accepted for publication at the \href{https://www.aclweb.org/portal/content/1st-personalization-generative-ai-personalize-workshop}{1st PERSONALIZE workshop} at EACL 2024
\end{textblock*}

\section{Introduction}

Natural language processing research has plunged headlong into the new alchemical science of \textit{prompt engineering} \cite{liuPretrainPromptPredict2023a}. Ask OpenAI's ChatGPT to ``think step-by-step'' and behold its improved reasoning performance \cite{weiChainofThoughtPromptingElicits2023}. Tell it to behave as an expert and witness its expertise increasing \cite{salewskiInContextImpersonationReveals}. 

The responsiveness of foundation models to prompt engineering has led researchers from diverse disciplines to explore their applications. This is certainly true in political science, where several recent studies investigate whether the malleability of LLMs would allow them to simulate the attitudes and behaviors of human subpopulations \citep{ chu_language_2023, jiang_communitylm_2022, kim_ai-augmented_2023,simmonsLargeLanguageModels2023,linegarLargeLanguageModels2023}. 

Polling plays an important role in opinion aggregation, acting as a cornerstone of governance \cite{shapiroPublicOpinionAmerican2011}. The use of LLMs as subgroup simulators has hypothesized benefits including decreased cost and increased sample sizes \cite{argyleOutOneMany2023}. As response rates to traditional survey methods decline, social scientists are encouraged to explore new methods \cite{ziemsCanLargeLanguage}. More than a dozen examples of the subpopulation representative modeling approach are found in academic research \cite{simmonsLargeLanguageModels2023}, and the approach has already garnered attention at local \cite{TalkCity} and national levels \cite{IONPrimulConsilier}.

\begin{figure}[t]
\centering
\includegraphics[width=0.9\columnwidth]{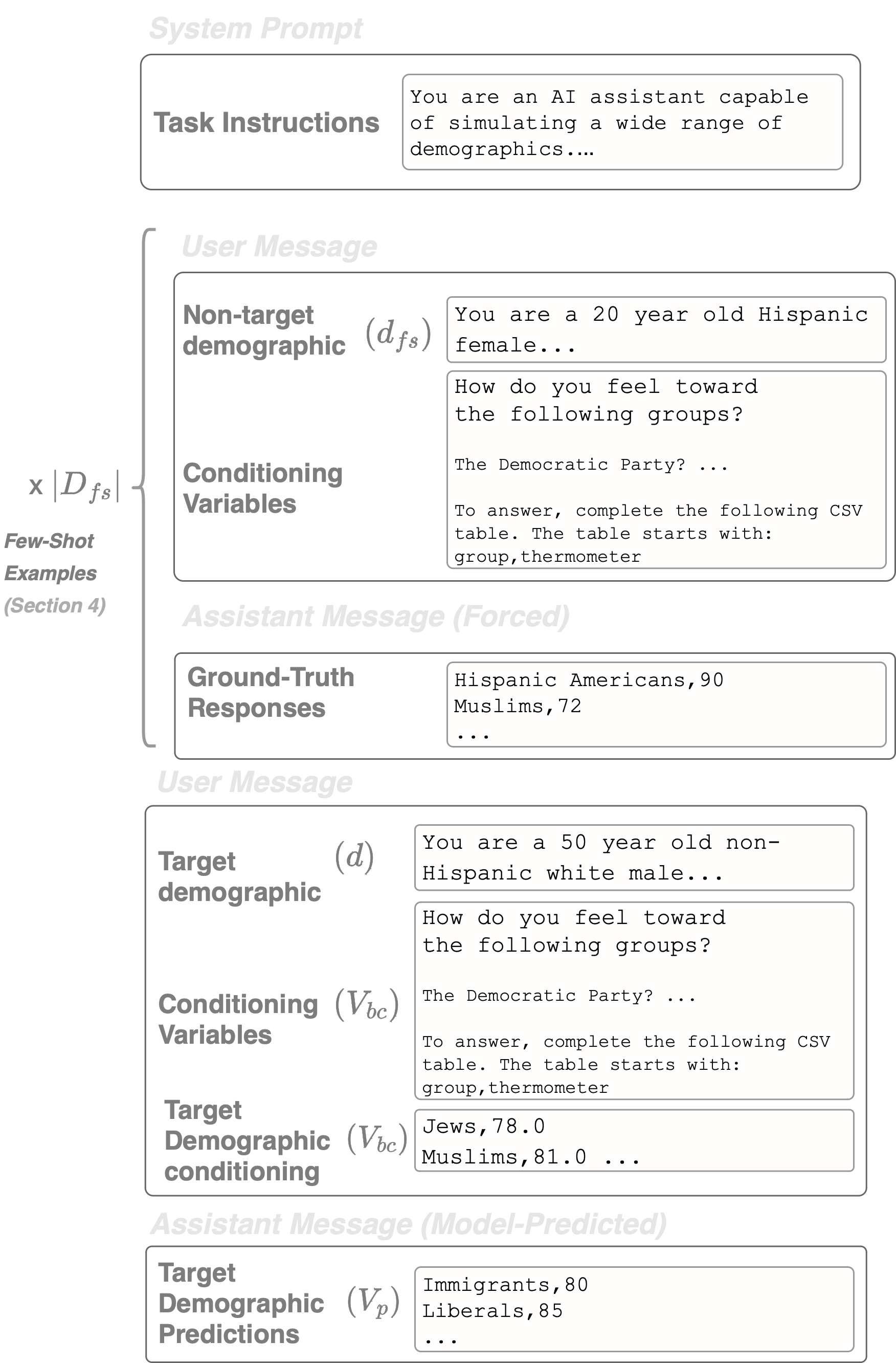} 
\caption{Description of a prompting strategy used for both RQ 1 and RQ 2. For Study 1, $|D_{f s}| = 0$.}
\label{fig:full-prompting-setup}
\end{figure}

\subsection{Limitations of LLMs as Subpopulation Representative Models}
While the potential benefits are considerable, applying LLMs as a substitute or complement for polling should be taken with caution. Recent work shows that prompting LLMs with demographic information leaves much to be desired. \citet{bisbeeArtificiallyPreciseExtremism2023} show that when ChatGPT (gpt-3.5-turbo)\footnote{https://openai.com/blog/chatgpt} is prompted with demographic information from the ANES Survey\footnote{The American National Election Studies (ANES) are national surveys of voters in the United States, conducted before and after presidential elections, with data since 1948 \cite{ANESHistory}} and asked to complete a Feeling Thermometer\footnote{The ANES Feeling Thermometer measures respondent affinity to various political groups in the United States.}, its responses are more extreme and less variable than the responses collected from human participants \citep{bisbeeArtificiallyPreciseExtremism2023}. \citet{santurkarWhoseOpinionsLanguage2023} show that overall fidelity\footnote{\citet{santurkarWhoseOpinionsLanguage2023} refer to this as alignment. We use ``fidelity" since alignment take several meanings.} to human response distributions is low for OpenAI's ChatGPT when the model is not prompted with a demographic descriptor, that prompting is more effective for some population subgroups than others, and that the fidelity obtained using demographic prompting, while higher than without, is still far from perfect. These studies express a pessimistic stance about the potential of LLMs for subpopulation representative modeling, one that is rightly held based on the experimental evidence to date. 

\subsection{In-Context Learning for Better Fidelity}
We contend here that we should not be prematurely pessimistic. As with chain-of-thought \cite{weiChainofThoughtPromptingElicits2023} and expert prompting, perhaps a straightforward technique to improve the performance of LLMs as subpopulation representative models has been overlooked. A hallmark of LLMs is their capability for in-context learning (ICL;  \citealt{brownLanguageModelsAre2020, dongSurveyIncontextLearning2023}). 
One popular mode of in-context learning is \textit{few-shot learning}, where task examples are provided in the context window to condition generation \cite{songComprehensiveSurveyFewshot2023}. Few-shot learning improves performance relative to prompting without examples, on tasks including translation and question answering \cite{brownLanguageModelsAre2020}, clinical information extraction \cite{agrawalLargeLanguageModels2022}, reading comprehension and natural language inference \cite{chowdheryPaLMScalingLanguage2022},
and improves factual accuracy of model responses \cite{semnaniWikiChatStoppingHallucination2023}.

The subpopulation representative modeling (SRM) task involves predicting the distribution of some response variables, such as candidate preference \cite{palakodetyMiningInsightsLargeScale2020}, feeling thermometer \cite{argyleOutOneMany2023} or stance on divisive issues \cite{kim_ai-augmented_2023} for a population subgroup (target demographic) identified by a combination of demographic variables. Applying LLMs to this task typically involves prompting the language model with a natural language description of the demographic and adding instructions to encourage the model to predict the response distribution. Importantly, this zero-shot approach does not leverage observed data from the subpopulation other than its demographic descriptors.

The most straightforward way to apply ICL to the SRM task would be to condition the model with data from the target subpopulation and demographic variables. With sufficient grounding in the target task, we expect that models could become representative. However, in this setup the practitioner has gained little, since they have to provide data from the target subpopulation and response variables to elicit desirable performance.\footnote{This setup could still be used for synthetic data generation for the subpopulation and response variables, similar to missing data imputation in \citet{kim_ai-augmented_2023}} For this reason, we expect that SRM practitioners would be enthusiastic to use available data to improve performance on unrelated subpopulations or unrelated response variables. In other words, \textit{generalization} beyond the data presented in the few-shot examples would allow practitioners to apply SRMs with improved performance even if data was not abundant for the subpopulation of interest.

\subsection{The Importance of Generalization}

For the subpopulation representative modeling task, generalization can occur along two axes: (1) generalization across response variables and (2) generalization across demographics. If a model can generalize across response variables, this means that conditioning on observed response variables improves fidelity for unobserved response variables. If a model can generalize across demographics, this means that conditioning on observed demographics improves fidelity for unobserved demographics. If these capabilities are demonstrated, the outlook for subpopulation representative modeling via LLMs may not be as dire as it seems. In-context learning could mitigate known issues such as extremism \cite{bisbeeArtificiallyPreciseExtremism2023} or lack of representativeness \cite{santurkarWhoseOpinionsLanguage2023}.

Successful generalization alone does not imply that LLMs are suitable for use as SRMs. However, we argue that if generalization were possible, it would encourage further development of LLM-based SRM technology. Integrating LLMs into the political infrastructure could have serious social consequences. For that reason, we believe that machine learning practitioners, social scientists, and policymakers should understand the viability of the technology, as greater viability may translate to an increased chance of real-world use. This motivates our study of the generalization capabilities of LLMs for the subpopulation representative modeling task.

\begin{table}[t]
\centering
\begin{tabular}{l|l}
\hline
\textbf{Demographic Variables} & \textbf{Response Variables} \\
\hline
age & The Democratic Party \\
race & The Republican Party \\
gender & Black Americans \\
income & White Americans \\
education & Hispanic Americans \\
political party & Asian Americans \\
 & Muslims \\
 & Christians \\
 & Jews \\
 & Liberals \\
 & Conservatives \\
\hline
\end{tabular}
\caption{Demographic and response variables used in this study.}
\label{table:appendix:vars}
\end{table}

\subsection*{Research Questions}

We address the following research questions:

\begin{itemize}
    
    \item \textbf{RQ 1 (Generalization across Response Variables)}: How does the fidelity of LLMs to some target \textbf{demographic} vary with the number of \textbf{response} variables from the target demographic used for conditioning? We address this in \Cref{sec:study1}.
    
    \item \textbf{RQ 2 (Generalization across Demographics)}: How does the fidelity of LLMs to some target \textbf{demographic} vary with the number of \textbf{examples from other demographics} used for conditioning? We investigate this in \Cref{sec:study2}.
\end{itemize}

\section{Methods}

This section documents methods shared across both studies. Specific methods for each study are documented in \Cref{sec:study1} and \Cref{sec:study2}.

\subsection{Data}

We use data from the American National Election Studies (ANES). We used the time series cumulative data file for the ANES Survey\footnote{available at \url{https://electionstudies.org/data-center/anes-time-series-cumulative-data-file/}}, which contains six demographic variables (age, race, gender, income, education, and political party), and 11 Feeling Thermometer variables shown in \Cref{table:appendix:vars}. The ANES Feeling Thermometer is a series of ratings questions where survey participants rate their affinity towards various political groups on a continuous scale from 0-100. Across all years, the ANES data contains 68,224 observations. We selected observations from the years 2016 and 2020, yielding 12,550 observations. After removing observations with missing values, the dataset used for experiments contained 4,397 observations. See \Cref{sec:appendix_data} for additional details on data processing steps applied before prompting.

\subsection{The Subpopulation Representative Modeling Task}

Subpopulation data consists of a number of observations of some set of variables $V$, with each observation corresponding to a single individual.
Often, this set of variables contains some subset $V_d \subset V$ that describe the demographic characteristics of each individual, and some other subset $V_b \subset V$ capturing individual behaviors or attitudes. At a high level, the goal for the SRM task might be to approximate the distribution of $V_b$ conditioned on $V_d$. However, it is equally likely that practitioners are interested in predicting a specific behavior and have some other behavioral data available for conditioning, requiring generalization across response variables. We investigate this setting in \Cref{sec:study1}. Additionally, practitioners may have some paired (demographic, behavior) data available for certain demographic cells and want to predict the behavior for other demographic cells. We investigate generalization across demographics in \Cref{sec:study2}.

\subsection{Measuring Fidelity Error}

\begin{figure*}
\begin{equation}
E(d, V_c, D_{fs}) = \frac{1}{|V_p|} \sum_{v_p \in V_p} \left| \hat{y}(d, V_c, v_p, D_{fs}) - y(d, v_p) \right|.
\label{eq:fidelity-zero-shot}
\end{equation}

\begin{equation}
E(d, n_c, n_{fs}) = \frac{1}{n_r} \sum_{ D_{fs} \sim \mathcal{D}(n_{fs}, d) } \frac{1}{|\mathcal{V}_{c}(n_c,d)|} \sum_{V_c \in \mathcal{V}_{c}(n_c, d)} E(d, V_c, D_{fs}).
\label{eq:fidelity-few-shot}
\end{equation}
\end{figure*}

We are interested in assessing how LLM fidelity to some target demographic varies with the amount of empirical data used to condition the model. We use the term \textit{fidelity error} ($E$) to refer to the gap between the LLM response and ground truth data observed from humans in the demographic of interest. In our setting, the behavioral variables $V_b$ are Feeling Thermometer ratings across 11 political groups. To explore generalization across response variables, we select some $V_{bc} \subset V_b$ to be used for conditioning. The LLM is tasked to predict the remaining variables $V_p = V_b \setminus V_{bc}$.

To obtain ground truth for $V_b$ at the demographic level, we obtain an average respondent profile for each demographic cell by calculating the mean responses for each of the 220 demographic cells in the ANES data.

We define fidelity error for some target demographic $d$ as the difference between the empirical mean and the LLM-predicted response, averaged over the Feeling Thermometer variables included in $V_p$. In general this error varies by the conditioning variables ($V_c$), see \Cref{eq:fidelity-zero-shot}. The term $y(d,v_p)$ is the empirical mean Feeling Thermometer for demographic $d$ towards group $v_p$. The term $\hat{y}(d, V_c, v_p, D_{fs})$ is the LLM-predicted Feeling Thermometer data for demographic $d$ towards group $v_p$, conditioned on variables $V_c$ and few-shot data $D_{fs}$. In other words, \Cref{eq:fidelity-zero-shot} describes the fidelity error of the model conditioned on a specific set of few-shot examples.

\Cref{eq:fidelity-few-shot} estimates the overall fidelity error of the model by sampling $n_r$ sets of few-shot examples from the observed data. In our experiments we used $n_r=5$. The term $\mathcal{V}_{c}(n_c, d)$ is the set of sets of conditioning variables having $|V_c| = n_c$ elements that are available for demographic $d$. The term $\mathcal{D}_{fs}(n_{fs}, d)$ is the set of sets of few-shot examples having $|D_{fs}| = n_{fs}$ elements that are available for demographic $d$. 

\subsection{Generating LLM Responses}

Our prompting strategy is briefly outlined here and in \Cref{fig:full-prompting-setup}. In this study, we utilize OpenAI's \texttt{gpt-3.5-turbo}, accessed via the API. We adapt a similar prompting strategy to \citet{bisbeeArtificiallyPreciseExtremism2023}, altering prompts to accommodate Research Questions 1 and 2. This approach comprises a consistent system prompt for directing the model's behavior and a variable user prompt, tailored for each research question. For RQ 1, each query features a \textit{single} user prompt with an incomplete Feeling Thermometer table. For RQ 2, we supply the model with multiple user prompts, each paired with an example model response which contains a Feeling Thermometer table with ground truth data. 

For a detailed view of our prompting setup, refer to the \Cref{sec:appendix_prompting}.

\section{Generalization of In-Context Learning Across Response Variables}
\label{sec:study1}

This study investigates the generalization of in-context learning across
response variables (Research Question 1). We are interested in finding out to what extent increasing the number of conditioning variables improves fidelity to unobserved response variables.

\subsection{Methods}

Each prompt includes all demographic variables, plus a subset of the behavioral variables $V_{bc} \subset V_b$. We are interested in relating the number of behavioral variables used for conditioning ($|V_{bc}|$) to the fidelity error. For each demographic cell, we compute the mean empirical response data. Then for each possible value of $|V_{bc}| \in [0,10]$, we randomly sample $n_r$ sets of conditioning variables. The empirical mean response data for these variables are presented in each prompt as a partially-completed Feeling Thermometer table in CSV format, as shown in \Cref{fig:appendix:response-var-conditioning}. The model then completes the remaining rows of the table. For each prompt, we parse the model-completed portion of the Feeling Thermometer table into CSV format. We then calculate the fidelity error for each prompt by comparing these responses to the empirical mean response data for variables $V_p$.

\subsection{Fidelity Error Decreases with Increasing Conditioning Variables}

\begin{figure*}[t]
\centering
\includegraphics[width=2.\columnwidth]{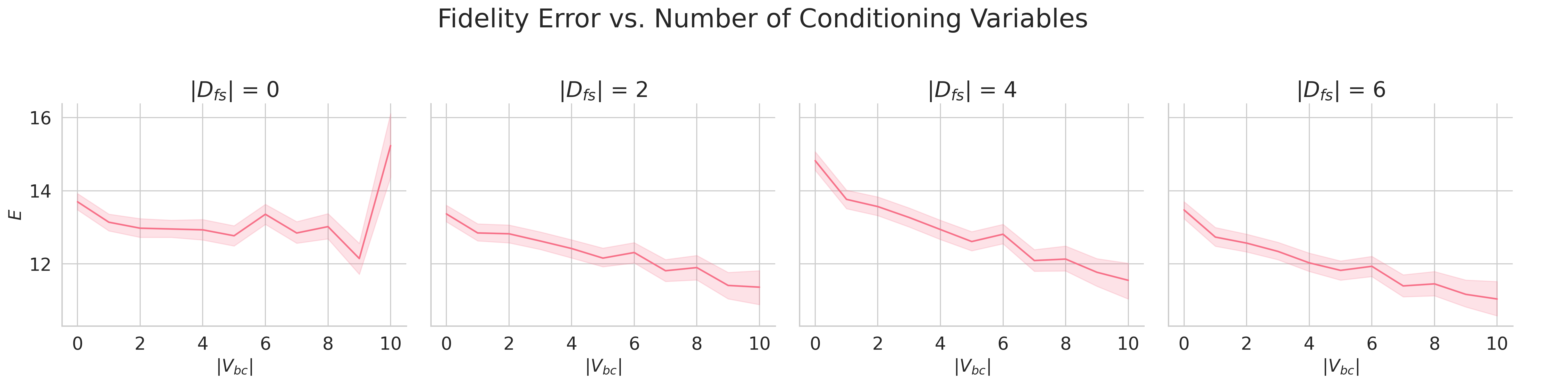}
\caption{Changes in the fidelity error depending on the $|V_{bc}|$ averaged across all demographics. The fidelity decreases as the number of conditioning variables increases. This pattern holds for every number of few-shot examples checked.}
\label{fig:num_groups:all}
\end{figure*}

\begin{figure*}[t]
\centering
\includegraphics[width=2.\columnwidth]{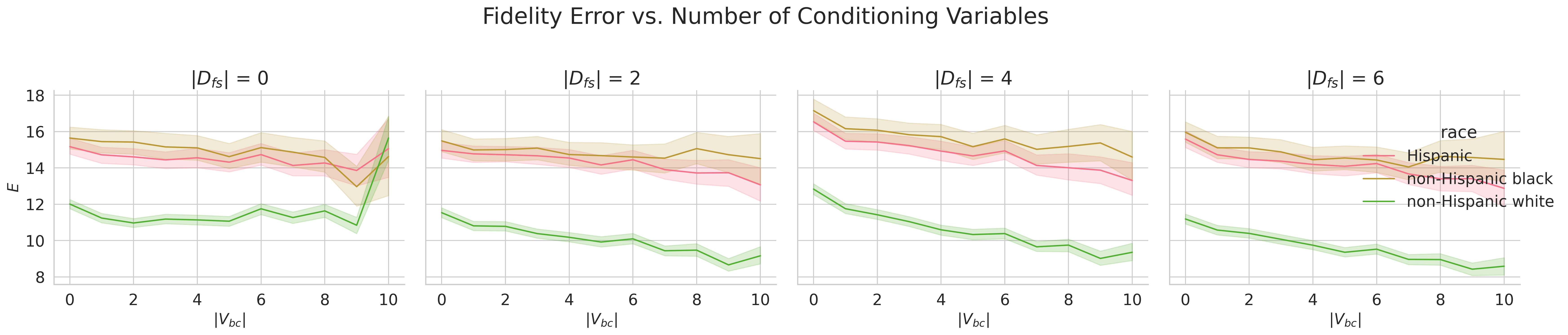}
\caption{Changes in the fidelity error ($E$) depending on the number of conditioning variables ($|V_{bc}|$) for different racial groups. Error rates are lower in general for non-Hispanic Whites than for other racial groups.}
\label{fig:num_groups:race}
\end{figure*}

\begin{figure*}[t]
\centering
\includegraphics[width=2.\columnwidth]{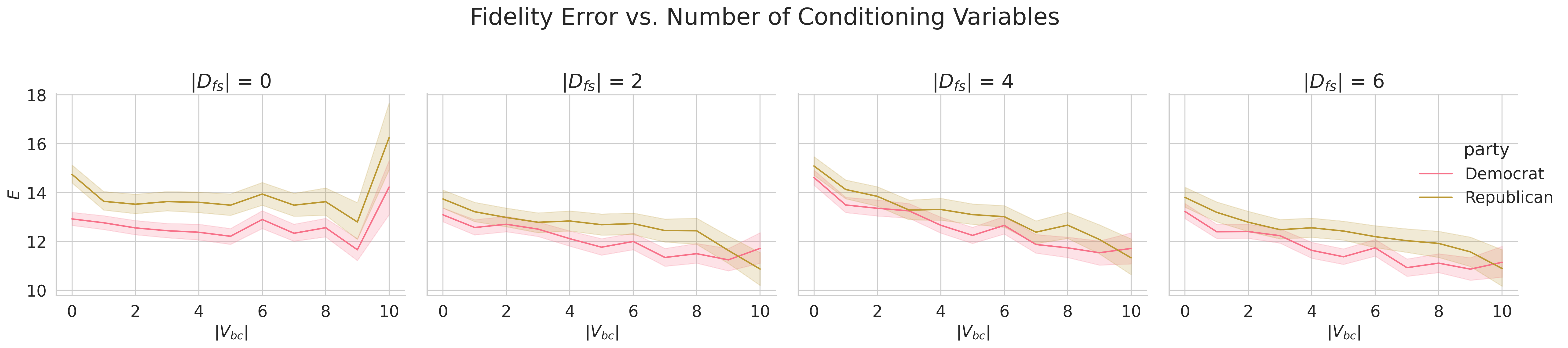}
\caption{Changes in the fidelity error ($E$) depending on the number of conditioning variables ($|V_{bc}|$) for different political parties. Error rates are lower in general for Democrats than for Republicans.}
\label{fig:num_groups:party}
\end{figure*}

\Cref{fig:num_groups:all} shows the relationship between number of conditioning variables $|V_{bc}|$ and the fidelity error for varying number of few-shot examples. In general, in-context conditioning on observed behavioral variables improves fidelity to unobserved behavioral variables, with error decreasing as the number of conditioning variables increases. 

\subsection{Effectiveness of Response Variable Conditioning Varies by Demographic}

We can observe discrepancies in error rates between demographics. For example, error rates are lower in general for non-Hispanic Whites than for other racial groups (\Cref{fig:num_groups:race}), and for Democratic party in comparison to the Republican one (\Cref{fig:num_groups:party}).

Reduction in error as a result of increased conditioning varies by demographic. For instance, error rates are roughly constant for $|V_{bc}| < 6$, then increase for the non-Hispanic black demographic, while continuing to decrease for the non-Hispanic white demographic (\Cref{fig:num_groups:race}). This suggests that conditioning on behavioral variables may be more effective for some demographics than for others.

Refer to \Cref{sec:appendix_figures} for figures showing relationships between the fidelity error and number of conditioning variables for other demographics.

\section{Generalization of In-Context Learning Across Demographics}\label{sec:study2}

\subsection{Methods}

In this study, we investigate the generalization of in-context learning across demographics.

In this case, we select some empirical data $D_{fs} \subset D, \{ d_{fs} \neq d \quad \forall \quad d_{fs} \in D_{fs} \}$ to be used as few-shot examples. Each prompt was constructed by selecting a target demographic, as in the previous study. Then $|D_{fs}| \in \{ 0, 2, 4, 6 \}$ few-shot examples of complete demographic and Feeling Thermometer information for non-target demographics were randomly selected. 

Selecting few-shot examples naturally raises the question of which examples to select. Few-shot example selection can be viewed as an information retrieval task, and many of the well-known methods from IR are applicable here. These include semantic similarity methods \cite{nanEnhancingFewshotTexttoSQL2023} as well as classic information retrieval algorithms such as max marginal relevance (MMR) \cite{carbonellUseMMRDiversitybased1998}. Few-shot example selection is also related to the problem of representative sampling in the social sciences \cite{manheimEmpiricalPoliticalAnalysis1981}; stratified sampling by demographic could be applied \cite{barretoBestPracticesCollecting2018}. Additionally, the recent trend towards larger models and LLM-as-a-service APIs has encouraged methods that maximize the number of few-shot examples to be included when the model input is restricted by total length \cite{SelectLengthLangchain}. 

However, the most straightforward approach is to sample uniformly at random from the observed data, and in this work we opt for this setup. Since the use of LLMs for SRM is relatively new and may be applied by practitioners who are not familiar with the aforementioned methods, we think it is important to consider the performance of naive methods. We are aware that the choice of sampling method could influence the results of this study; see the Discussion for commentary on the effects of sampling strategy and our suggestions for additional experiments.

\subsection{Fidelity Error Decreases with Increasing Few-Shot Examples}

\begin{figure*}[t]
\centering
\includegraphics[width=2.\columnwidth]{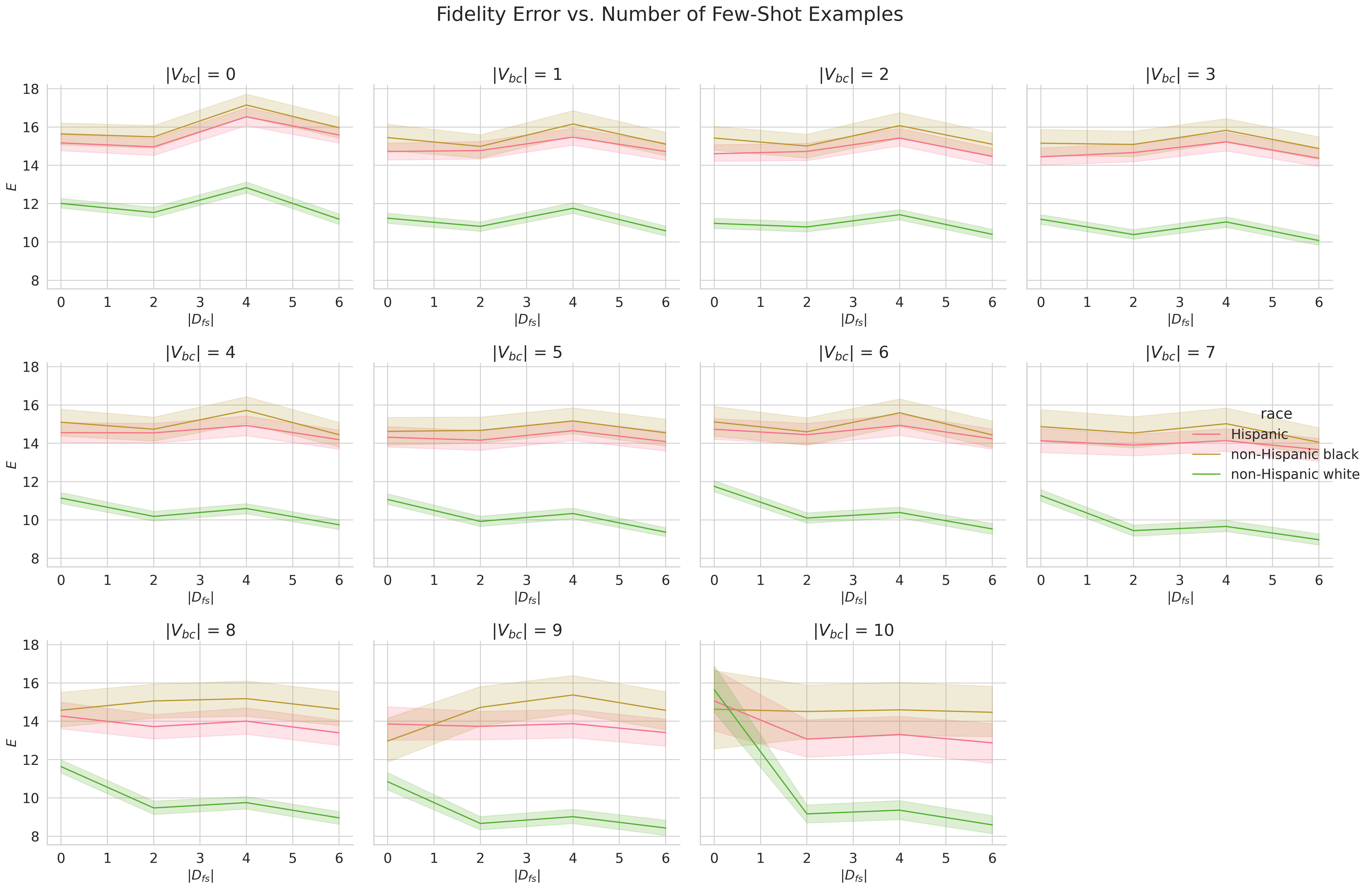}
\caption{Changes in the fidelity error ($E$) depending on the number of few-shot examples ($|D_{fs}|$) for different racial groups. Error rates are lower for non-Hispanic Whites. While with increased number of few-shot examples the fidelity error for other race groups remain nearly constant, the fidelity rate for non-Hispanic white racial group decreases.}
\label{fig:few_shot:race}
\end{figure*}

\begin{figure}[t]
\centering
\includegraphics[width=0.95\columnwidth]{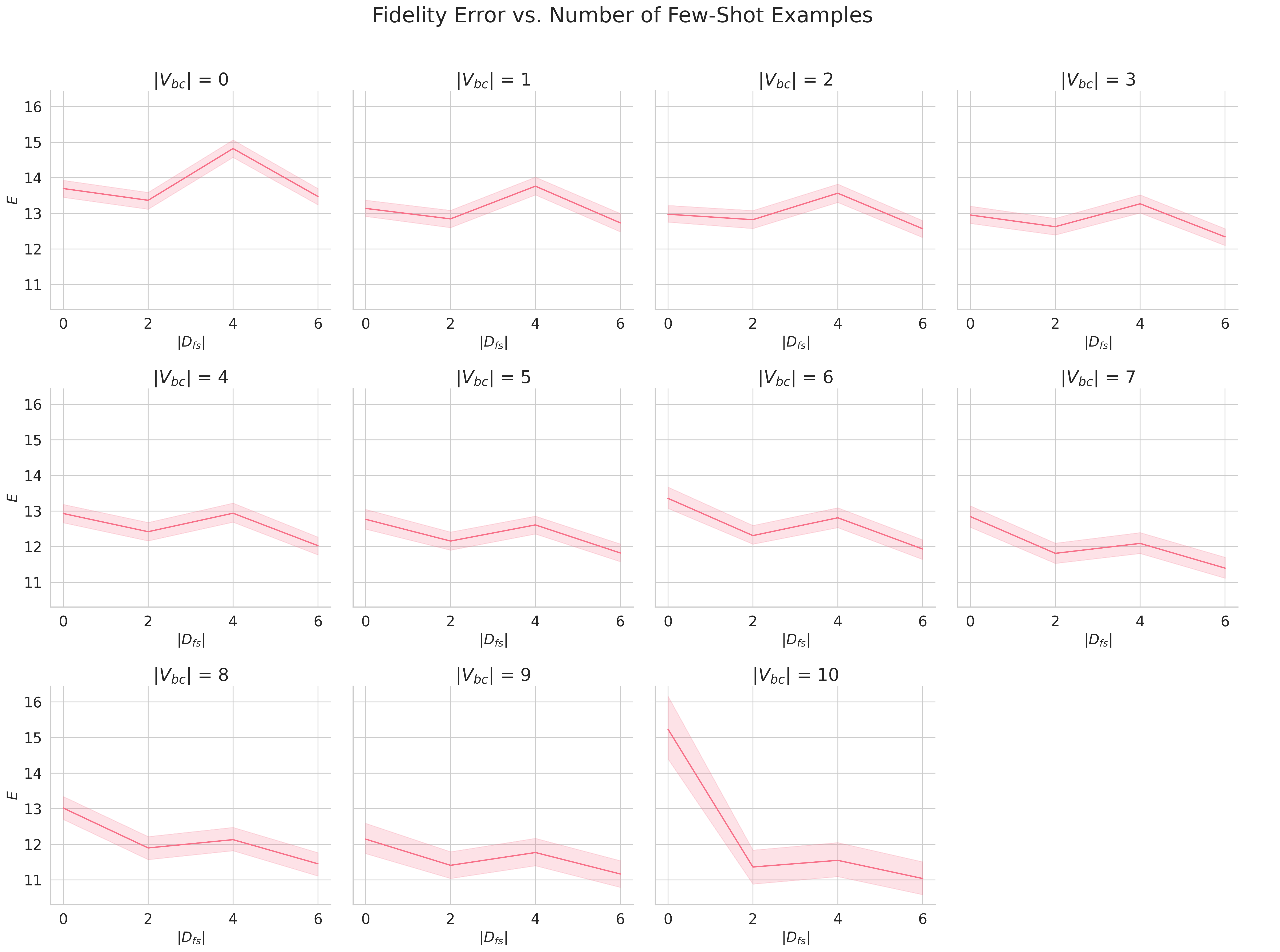}
\caption{Changes in the fidelity error depending on the $|D_{fs}|$ averaged across all demographics. The fidelity decreases as the number of few-shot examples increases. This pattern holds for every number of conditioning variables checked.}
\label{fig:few_shot:all}
\end{figure}

\Cref{fig:few_shot:all} shows fidelity error as it relates to the number of few-shot examples. In general, fidelity error decreases with increasing number of few-shot examples.

\subsection{Effectiveness of Few-Shot Learning Varies by Demographic}

We can again observe discrepancies in error rates between various demographics, including race, age income, and party. For instance, from \Cref{fig:few_shot:race} it can be seen that not only the fidelity error for non-Hispanic whites is smaller in general, but also that in-context learning is more efficient for this ethnicity.

We draw heavily on the prompting methods used in \citet{bisbeeArtificiallyPreciseExtremism2023} - this was done intentionally, for the sake of comparison. The key difference is the use of conditioning based on ground-truth data. \citet{bisbeeArtificiallyPreciseExtremism2023}'s study is one of the sharpest criticisms of LLM-based SRMs to date and raises important questions about the viability of LLM-based SRMs. If the deficiencies highlighted in this work are ameliorated by in-context learning, this would be an important consideration. We use similar methods so that results are attributable to the use of in-context learning, rather than differences in prompting strategy.

\section{Discussion: Subpopulation Representative Modeling via In-Context Learning}

Recent criticisms have argued that Large Language Models do not sufficiently represent the opinions or behaviors of human subpopulations when these subpopulations are specified in the context \cite{santurkarWhoseOpinionsLanguage2023,bisbeeArtificiallyPreciseExtremism2023}. However, extant work neglects the capability for models to learn via in-context learning.

Our experiments demonstrate that LLMs \textit{can} learn the subpopulation representative modeling task in-context. The experiments in \Cref{sec:study1} show that providing the model with partial information about subpopulation behavior improves model fidelity on unobserved response variables. \Cref{sec:study2} shows that providing the model with information about other subpopulations can improve model fidelity to an unrelated subpopulation of interest. 

In this experiment, we selected few-shot examples uniformly at random. This is only one of several few-shot example selection strategies available to the practitioner (see \Cref{sec:study2}, Methods). We believe it is likely that the example selection strategy has some influence over the performance disparities between majority and minority groups. Appendix \Cref{fig:appendix:non_nans_count} shows that the ANES data is imbalanced with repsect to the demographic variables -- for example, approximately three fourths of respondents were non-Hispanic white, as opposed to Hispanic or non-Hispanic black. The minority categories account for approximately 1/8th of the observations each. For a given target example, the likelihood to select a few-shot example with the same race is proportional to the distribution of the data over the race variable. In general, it is more likely that a randomly-selected few-shot example will share demographic values with the target example when the target example belongs to the majority demographic. Assuming that the similarity between few-shot examples correlates to their utility for the predictive task, this dataset bias could result in few-shot prompting being more effective for majority groups. This applies both in absolute terms, and in terms of the marginal benefit of additional few-shot examples. We encourage further investigation of the relationship between demographic representation in the few-shot data, performance discrepancies across demographics, and few-shot example selection strategies, and plan to explore this theme in future work.

These aggregate results seem promising for the potential of LLMs to perform the subpopulation modeling task. However, upon closer analysis, we find that the effectiveness of in-context learning is variable across demographics. While additional conditioning data boosts performance for some demographics, it has negligible or even deleterious effects for others. This result extends prior work showing variation across demographics in the exaggeration of stereotypical response patterns \citep{bisbeeArtificiallyPreciseExtremism2023} and the fidelity of LLMs to human responses without conditioning \citep{santurkarWhoseOpinionsLanguage2023}. The subgroup-specific effectiveness of in-context learning for SRM presents challenges for SRM practitioners, as well as decision-makers using the results of SRMs. We suggest three directions for future work. The inequitable performance of LLMs on subpopulation simulation calls the ethicality of the endeavor into question. In tasks like recidivism prediction, theoretical results indicate mutual unsatisfiability of model bias criteria \cite{kleinbergInherentTradeOffsFair2016,chouldechovaFairPredictionDisparate2017}. These impossibility results influence why the field views machine learning models as appropriate for certain use cases and possibly unfit for others. We encourage 
similar investigation into the ethical nature of the subpopulation representative modeling task. This should take into consideration the dual-use nature of subpopulation representative models -- that they could be leveraged for positive use cases (improving existing political representation processes) as well as negative (used to steer misinformation campaigns). 
Secondly, our results highlight the need for fine-grained benchmarking for subpopulation representative models, in terms of generalization performance in few-shot settings as studied here, as well as absolute performance in zero-shot settings. Finally, we note that several approaches have been proposed to ameliorate issues with existing subpopulation representation techniques \cite{santurkarWhoseOpinionsLanguage2023,lahoti_improving_2023-1}. We see potential for further research in this area of improving subpopulation representative model performance.

\bibliography{acl_latex}

\appendix

\begin{figure}[t]
\centering
\includegraphics[width=0.9\columnwidth]{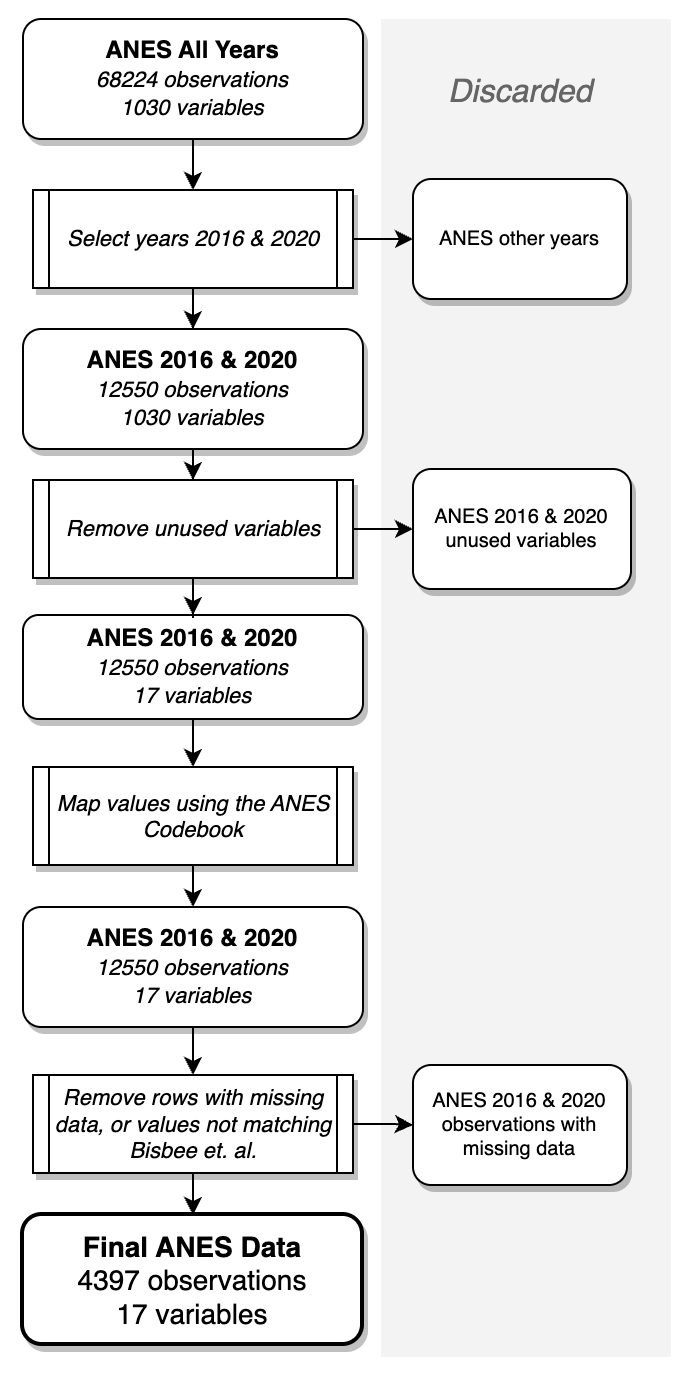} 
\caption{Data processing pipeline.}
\label{fig:appendix:data_pipeline}
\end{figure}

\section{Prompting Setup}\label{sec:appendix_prompting}

This section provides specific prompts used for our study, further explaining \Cref{fig:full-prompting-setup}. Our strategy is adapted from the methodology used in \citet{bisbeeArtificiallyPreciseExtremism2023}, with several modifications to better suit our research objectives.

The prompting setup consists of task instructions (the system prompt) and the user prompt. The system prompt is a constant element in all the requests, designed to guide the AI model towards displaying a subpopulation representative behavior (see \Cref{fig:appendix:system-prompt} for the exact structure of the system prompt). User prompts vary based on the research question. For research question 1, a single user prompt is used. This prompt consists of three parts:
\begin{enumerate}
    \item Target demographic description ($d$), which provides demographic data to the model (see \Cref{fig:appendix:user-message});
    \item A number of conditioning variables, constant across all requests, posing a concrete question with respect to the current study (see \Cref{fig:appendix:user-prompt});
    \item Target demographic conditioning, providing a partial Feeling Thermometer table of $V_{bc} \subset V_b$ variables into the model (see \Cref{fig:appendix:response-var-conditioning}).
\end{enumerate}

\begin{figure}[t]
\centering
\begin{scriptsize}
\begin{lstlisting}[numbers=none]
The table starts with: 
group,thermometer
Muslims,30.0
Jews,72.0
\end{lstlisting}
\end{scriptsize}
\caption{Possible end of the table to condition on two response variables for the target demographic (ground-truth responses).}
\label{fig:appendix:response-var-conditioning}
\end{figure}

For research question 2, multiple user prompts are employed, each paired with an example model response. The structure of additional user prompts remains consistent with that of the RQ 1, but instead of the target demographic data, a non-target demographic conditioning is used. All but one of these prompts serve as few-shot examples. In RQ 2, a single few-shot example consists of:
\begin{enumerate}
    \item \textbf{Non}-target demographic data ($d_{fs}$), fed into the model in the same format as in the \Cref{fig:appendix:user-message};
    \item A number of conditioning variables, similarly to RQ 1 (refer to \Cref{fig:appendix:user-prompt});
    \item Ground-truth responses from the cleaned ANES data -- a Feeling Thermometer table for the same non-target demographic. The format is the same as in \Cref{fig:appendix:response-var-conditioning}.
\end{enumerate}

Few-shot examples are followed by the final user prompt (same as in the RQ 1), soliciting a model prediction for the feeling thermometer table for the target demographic.

\begin{figure*}[t]
\centering
\begin{scriptsize}
\begin{lstlisting}[numbers=none]
You are an AI assistant capable of simulating a wide range of demographics.

You receive demographic data describing a single person as an input.
Provide responses from this person's perspective.
Use only knowledge about politics that they would have.

The user asks about individuals' feelings toward different groups.
Responses should be given on a scale from 0 (meaning cold feelings) to 100 (meaning warm feelings).
Ratings between 50 degrees and 100 degrees mean that you feel favorable and warm toward the group. 
Ratings between 0 degrees and 50 degrees mean that you don't feel favorable toward the group and that you don't care too much for that group.
You would rate the group at the 50 degree mark if you don't feel particularly warm or cold toward the group.
\end{lstlisting}
\end{scriptsize}
\caption{System prompt used for all studies (task instructions).}
\label{fig:appendix:system-prompt}
\end{figure*}

\begin{figure*}[t]
\centering
\begin{scriptsize}
\begin{lstlisting}[numbers=none]
You are a {age} year old {race} {gender} with a {education}, earning ${income} per year.
You are a registered {party} living in the USA in 2019.
\end{lstlisting}
\end{scriptsize}
\caption{User message, used either for non-target ($d_{fs}$) or target demographic $d$.}
\label{fig:appendix:user-message}
\end{figure*}

\begin{figure*}[t]
\centering
\begin{scriptsize}
\begin{lstlisting}[numbers=none]
How do you feel toward the following groups?
The Democratic Party? 
The Republican Party? 
Black Americans? 
White Americans? 
Hispanic Americans? 
Asian Americans? 
Muslims?
Christians?
Jews?
Liberals?
Conservatives?

To answer, complete the following CSV table.

The table starts with:
group,thermometer
\end{lstlisting}
\end{scriptsize}
\caption{Conditioning variables ($V_{bc}$).}
\label{fig:appendix:user-prompt}
\end{figure*}

\section{Notes on Data Processing}
\label{sec:appendix_data}

\begin{figure*}[ht!]
\centering
\includegraphics[width=2.\columnwidth]{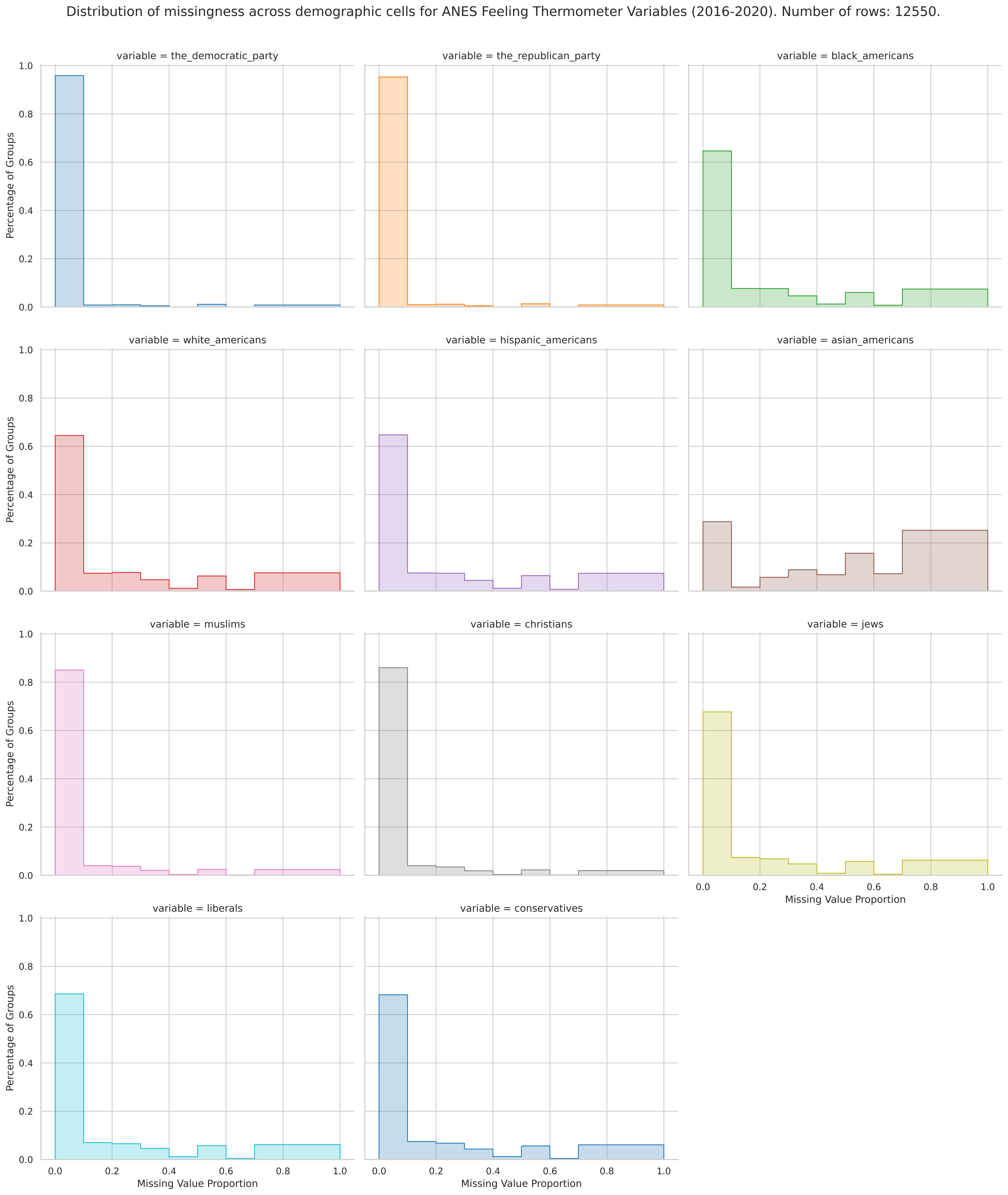} 
\caption{Distribution of missingness across demographic cells for ANES feeling thermometer Variables (2016-2020).}
\label{fig:appendix:missingness_distribution}
\end{figure*}

\begin{figure*}[ht!]
\centering
\includegraphics[width=0.95\columnwidth]{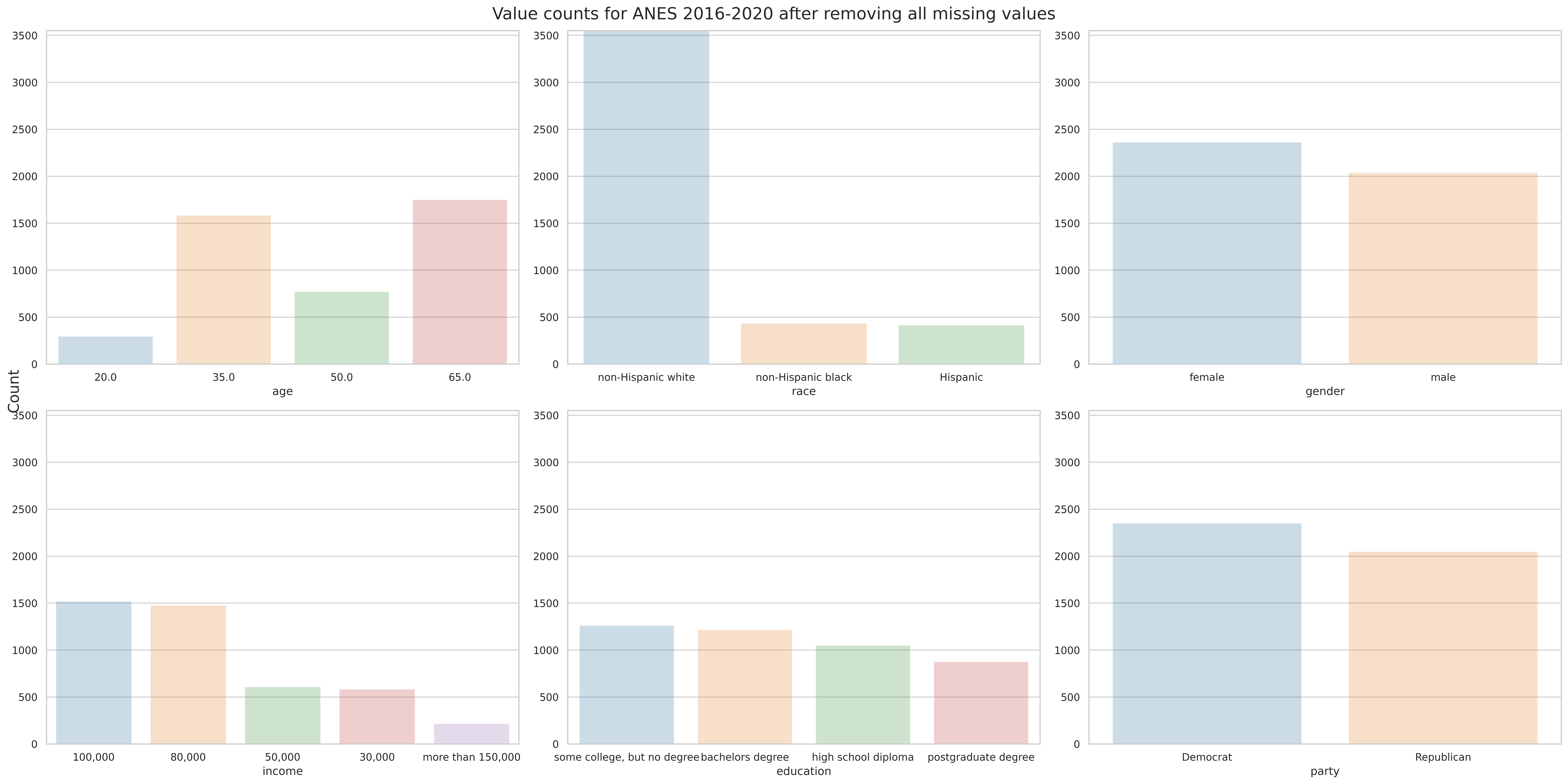} 
\caption{Value counts for ANES 2016-2020 after removing all missing values.}
\label{fig:appendix:non_nans_count}
\end{figure*}

Many observations in the ANES data were incomplete. Missing value rates for the data are shown in \Cref{fig:appendix:missingness_distribution}. Counts of observations for each demographic variable are shown in \Cref{fig:appendix:non_nans_count}. Removing rows containing missing demographic and response variables, and observations with variable values other than those in \Cref{table:appendix:vars} resulted in 4,397 observations, with 570 unique demographic cells. The end-to-end data processing pipeline is shown in \Cref{fig:appendix:data_pipeline}.

\section{Figures}\label{sec:appendix_figures}

Figures \ref{fig:appendix:few_shot_age}, \ref{fig:appendix:num_groups_age}, \ref{fig:appendix:few_shot_education}, \ref{fig:appendix:num_groups_education}, \ref{fig:appendix:few_shot_gender}, \ref{fig:appendix:num_groups_gender}, \ref{fig:appendix:few_shot_income}, \ref{fig:appendix:num_groups_income}, \ref{fig:appendix:few_shot_party}, \ref{fig:appendix:num_groups_party}, \ref{fig:appendix:few_shot_race} and \ref{fig:appendix:num_groups_race} show our findings for different demographic variables.

\foreach \x in {age, education, gender, income, party, race} {
  \begin{figure*}[t]
    \centering
    \includegraphics[width=2.\columnwidth]{figures/fidelity_error_vs_few_shot_k_\x_wrapped.png}
    \caption{Fidelity error vs. $|D_{fs}|$ for \x.}
    \label{fig:appendix:few_shot_\x}
  \end{figure*}
  
  \begin{figure*}[t]
    \centering
    \includegraphics[width=2.\columnwidth]{figures/fidelity_error_vs_num_groups_\x_wrapped.png}
    \caption{Fidelity error vs. $|V_{bc}|$ for \x.}
    \label{fig:appendix:num_groups_\x}
  \end{figure*}
}

\end{document}